%% file: main.tex
\documentclass[10pt,twocolumn,letterpaper]{article}

\usepackage{cvpr}              

\input{preamble}

\usepackage{soul}

\definecolor{cvprblue}{rgb}{0.21,0.49,0.74}
\usepackage[pagebackref,breaklinks,colorlinks,citecolor=cvprblue]{hyperref}

\newcommand{\cuthalfcaptionup}{\vspace*{-5pt}}
\newcommand{\ignore}[1]{}

\title{SCALAR-NeRF: SCAlable LARge-scale Neural Radiance Fields for \\ Scene Reconstruction}

\author{
Yu Chen \qquad\qquad\qquad Gim Hee Lee\\
Department of Computer Science, National University of Singapore\\
{\tt\small chenyu@comp.nus.edu.sg}\qquad {\tt\small gimhee.lee@nus.edu.sg} \\
}

\begin{document}
\twocolumn[{%
\renewcommand\twocolumn[1][]{#1}%
\maketitle
\vspace{-18pt}
\input{fig_tex/teaser}
}]

\input{sec/0_abstract}    
\input{sec/1_intro}
\input{sec/2_related_work}
\input{sec/3_method}
\input{sec/4_experiments}

\input{sec/5_conclusion}
\newpage
{
    \small
    \bibliographystyle{ieeenat_fullname}
    \bibliography{main}
}

\input{sec/X_suppl}

\end{document}

%% file: preamble.tex
\usepackage{times}
\usepackage{epsfig}
\usepackage{graphicx}
\usepackage{amsmath}
\usepackage{amssymb}
\usepackage{booktabs}
\usepackage{caption}
\usepackage{multirow}
\usepackage{makecell}
\usepackage{color}
\usepackage{algorithm}
\usepackage[noend]{algpseudocode}
\usepackage{subcaption}
\usepackage{bbding}
\usepackage{soul}
\usepackage{float}
\usepackage{setspace}

\usepackage[dvipsnames]{xcolor}

\makeatletter
\@namedef{ver@everyshi.sty}{}
\makeatother
\usepackage{tikz}
\usepackage{colortbl} 

\definecolor{bronze}{RGB}{205, 127, 50}
\definecolor{silver}{RGB}{192,192,192}
\definecolor{gold}{HTML}{FFD700}

\newcommand{\mycircle}[1][gold]{
  \begin{tikzpicture}
    \draw[#1, fill=#1] (0, 0) circle (0.12cm);
  \end{tikzpicture}
}


%% file: fig_tex/teaser.tex
\begin{center}
    \captionsetup{type=figure}
    \cuthalfcaptionup
    \centering
    \includegraphics[width=1.\linewidth]{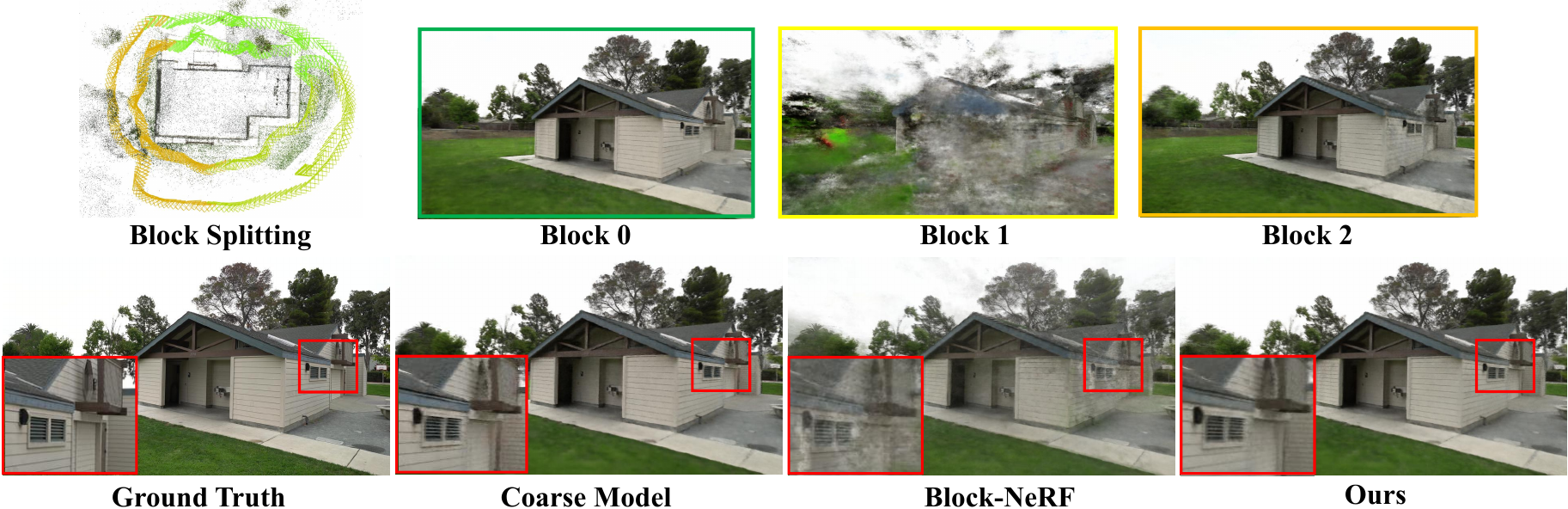}
    \vspace{-4mm}
  
    \captionof{figure}{\textbf{Comparison of our method to Block-NeRF}.
    Our method adopts a coarse-to-fine framework to train NeRF models under large-scale scenes, which effectively 
    avoids the `foggy' issue in Block-NeRF.
    }
    \cuthalfcaptionup
    \label{fig:teaser}
\end{center}%

%% file: sec/0_abstract.tex
\begin{abstract}
\vspace{-2.5mm}
In this work, we introduce \textbf{SCALAR-NeRF}, a novel 
framework tailored for scalable large-scale neural scene reconstruction. 
We structure the neural representation as an encoder-decoder architecture, where the encoder processes 3D point coordinates to produce encoded features, and the decoder generates geometric values that include volume densities of signed distances and colors.
Our approach first trains a coarse global model on the entire image dataset. 
Subsequently, we partition the images into smaller blocks using KMeans 
with each block being modeled by a dedicated local model. We enhance the overlapping regions across different blocks by scaling up the bounding boxes of each local block. 
Notably, the decoder from the global model is shared across distinct blocks 
and therefore promoting alignment in the feature space of local encoders.
We propose an effective and efficient methodology to fuse the outputs from these local models to attain the final reconstruction. Employing this refined coarse-to-fine strategy, our method outperforms state-of-the-art NeRF methods and 
demonstrates scalability for large-scale scene reconstruction. The code will be available on our \href{https://aibluefisher.github.io/SCALAR-NeRF/}{project page}.
\end{abstract}

%% file: sec/1_intro.tex
\section{Introduction}
\label{sec:intro}
Over the past three decades, 3D scene reconstruction remains an active research field with useful applications spanning augmented reality, 
real-world simulation, 
localization, \textit{etc}. The advent of neural scene reconstruction techniques
~\cite{DBLP:conf/eccv/MildenhallSTBRN20,DBLP:conf/nips/WangLLTKW21} has revolutionized the digitalization 
of 3D scenes, unveiling a realm beyond the limitations of traditional photogrammetry tools~\cite{DBLP:conf/cvpr/SchonbergerF16}. 
These neural methods exhibit superior resilience to non-Lambertian effects, appearance changes~\cite{DBLP:conf/cvpr/Martin-BruallaR21}, 
and dynamic scenes~\cite{DBLP:conf/cvpr/PumarolaCPM21} compared to their traditional counterparts.

Recent advancements have significantly improved the efficiency of training neural scene representations. While earlier approaches require long training time on a vanilla NeRF~\cite{DBLP:conf/eccv/MildenhallSTBRN20}, innovative techniques such as 
Instant-NGP~\cite{DBLP:journals/tog/MullerESK22} and TensoRF~\cite{DBLP:conf/eccv/ChenXGYS22} have remarkably condensed this 
training time to mere minutes. Nonetheless, the majority of these developments primarily focus on small-scale object-centric 
scenes, leaving the frontier of neural reconstruction on large-scale scenes largely unexplored.

The pursuit of neural reconstruction in large-scale scenes is not an easy endeavour, where significant challenges arise from several factors:
\begin{itemize}
\item \textbf{Parameter Overload}. The sheer volume of parameters required for updating poses a significant obstacle. Although augmenting model 
parameters might enhance reconstruction quality, it becomes challenging to train and manage such large models on a single GPU.

\item \textbf{Extended Training Duration}. Larger model capacities demand extended training times to capture high-fidelity scene details, thereby 
posing a time-intensive challenge.

\item \textbf{Incomplete Scene Details}. Mere augmentation of model parameters may still result in missing scene details, highlighting the 
limitations of this simplistic approach.

\end{itemize}

To overcome inherent limitations and extend neural scene reconstruction to city-scale scenes, recent works such as 
Block-NeRF~\cite{DBLP:conf/cvpr/TancikCYPMSBK22} and Mega-NeRF~\cite{DBLP:conf/cvpr/TurkiRS22} have embraced a divide-and-conquer 
approach. While Block-NeRF adeptly addresses the challenge of maintaining appearance consistency across different blocks, noteworthy 
challenges persist, particularly when block centroids are in close proximity. In such cases, Block-NeRF tends to generate images 
with a foggy appearance, attributed to its straightforward inverse distance weighting function for image blending. On the other 
hand, Mega-NeRF primarily focuses on utilizing aerial images for applications such as rescue searches, presenting its unique set 
of constraints and limitations.

In this paper, we introduce a novel coarse-to-fine framework for reconstructing large-scale scenes using NeRF. Our method leverages 
a multi-resolution hashing table as the feature encoder. Although the multi-resolution hash table is widely adopted in neural scene 
reconstruction, we observe that simply increasing the hash table size not only slows down training but also renders it infeasible 
to fit into a single GPU during training.

The core insight of our method is that a large hash table cannot be accommodated by a non-commercial GPU, while a small hash table 
has a higher probability of hash collisions. Splitting scenes into smaller blocks enables training local models with finer details 
without the necessity of increasing the hash table size. Initially, we train a rapid, coarse global model with all training images. 
Despite its expedited training, this coarse model retains considerable reconstruction quality. Subsequently, we partition the 
training images into smaller blocks, each reconstructed using an individual local model. We further enhance the intersection areas 
of local blocks by expanding the bounding boxes that cover the images in the current block. 
To improve the regularization of the hash encoder in each local model, we share the decoder from the coarse model across all local 
models. The shared decoder can either be fixed for faster training speed or fine-tuned for better reconstruction quality. 
Unlike Block-NeRF, which blends images using the inverse distance weighting function, we synthesize images from each block and fuse 
them with a reference image rendered from the coarse model. 
Leveraging the \textbf{\textit{coarse-global to fine-local}} strategy, our method with a hash table size of $2^{19}$ surpasses Block-NeRF and Instant-NGP with a hash table size of $2^{21}$.

We train and evaluate our model on outdoor large-scale scenes. Experiments demonstrate its substantial performance, surpassing 
existing state-of-the-art methods such as Block-NeRF. The improvements are evident in the scene reconstruction showcased 
in Fig.~\ref{fig:teaser}. The main contributions of our work are:
\begin{itemize}
\item We propose a novel framework for reconstructing large-scale scenes.
\item The proposed method enables training NeRF models under large-scale scenes on a single GPU without 
compromising reconstruction quality.
\item Extensive experiments on the large-scale Tanks and Temples dataset show that our method surpasses existing NeRF methods.
\end{itemize}

%% file: sec/2_related_work.tex
\section{Related Work}
\label{sec:related_work}

\paragraph{Large-Scale 3D Reconstruction.} Snavely \etal~\cite{DBLP:journals/tog/SnavelySS06} pioneered a system enabling virtual 
visits to global attractions. They utilized Structure-from-Motion (SfM)~\cite{DBLP:conf/cvpr/SchonbergerF16} and keypoint extraction 
with SIFT~\cite{DBLP:journals/ijcv/Lowe04} to reconstruct sparse scene structures. Image similarity searching methods like vocabulary trees~\cite{DBLP:conf/cvpr/NisterS06} 
and NetVLAD~\cite{DBLP:conf/cvpr/ArandjelovicGTP16} were applied to enhance reconstruction efficiency by removing unnecessary matching 
pairs. The works by Zhu \etal~\cite{Zhu2017Parallel,DBLP:conf/cvpr/ZhuZZSFTQ18} and Chen \etal~\cite{DBLP:journals/pr/ChenSCW20,DBLP:conf/icra/ChenYSYLL23} adopted a 
divide-and-conquer strategy 
for scalability under large-scale scenes. The improvement of reconstruction quality and efficiency in large-scale scenes has been 
significantly influenced by Multi-View Stereo (MVS) techniques. To recover the scene details as much as possible, MVS
~\cite{DBLP:journals/ftcgv/FurukawaH15,DBLP:conf/cvpr/FurukawaCSS10} is used to densify scene structures by plane 
sweep~\cite{DBLP:conf/cvpr/GallupFMYP07} or patch-match~\cite{DBLP:journals/tog/BarnesSFG09}. Recently, Yao 
\etal~\cite{DBLP:conf/eccv/YaoLLFQ18} leverages 3D convolutions to learn the depth probability map from the cost volume constructed 
by plane sweep. Later works~\cite{DBLP:conf/iccv/LuoGJHL19,DBLP:conf/cvpr/0008LLSFQ19,DBLP:conf/cvpr/WangGVSP21,DBLP:conf/cvpr/WangGVP22} 
also follow a similar paradigm to improve the reconstruction quality and efficiency of MVS.

\vspace{-4mm}
\paragraph{Novel View Synthesis.} The field of novel view synthesis has seen significant advancements, notably with neural radiance 
fields~\cite{DBLP:conf/eccv/MildenhallSTBRN20} enabling rendering from novel viewpoints with encoded frequency 
features~\cite{DBLP:conf/nips/TancikSMFRSRBN20}. To improve the rendering 
efficiency, NSVF~\cite{DBLP:conf/nips/LiuGLCT20} represents scenes into sparse voxels, where the voxels are gradually updated by checking 
the volume densities. PlenOctrees~\cite{DBLP:conf/iccv/YuLT0NK21} also uses sparse voxels for scene representation. To improve the training 
speed and rendering efficiency, the multi-layer perceptron (MLP) is converted to volume densities and spherical harmonics that are stored 
on the octree leaves. Plenoxels~\cite{DBLP:conf/cvpr/Fridovich-KeilY22} found that the MLP is not necessary, and instead uses tri-linear 
interpolation to improve the continuities of the scalar values stored on the voxel corners. Instant-NGP~\cite{DBLP:journals/tog/MullerESK22} 
encoded the features into a multi-resolution hash table, where the hash collision is implicitly handled during optimization. 
TensoRF~\cite{DBLP:conf/eccv/ChenXGYS22} uses CP-decomposition or VM-decomposition to encode scenes into three orthogonal axes and 
planes. The low-rank decomposition results in highly compact representations. Gaussian Splatting~\cite{DBLP:journals/tog/KerblKLD23} 
initializes 3D Gaussians from a set of sparse point clouds, the 3D Gaussians are used as explicit scene representation and dynamically 
merged and split during training. The splatting operation can be performed very fast through rasterization, which enables real-time 
novel view synthesis. Other methods also focus on the generalizability of NeRF
~\cite{DBLP:conf/cvpr/WangWGSZBMSF21,DBLP:conf/iccv/ChenXZZXY021,DBLP:conf/cvpr/JohariLF22,DBLP:conf/cvpr/SuhailESM22,DBLP:conf/cvpr/LiuPLWWTZW22}, 
bundle-adjusting camera poses and NeRF~\cite{DBLP:conf/iccv/LinM0L21,DBLP:conf/iccv/MengCLW0X0Y21,chen2023dbarf},
removing floaters~\cite{Sabour_2023_CVPR,DBLP:journals/corr/abs-2309-03185}, and leveraging sparse or dense depth to 
supervise the training of NeRF~\cite{DBLP:conf/iccv/WeiLRZL021,kangle2021dsnerf,roessle2022depthpriorsnerf}, \etc.

\vspace{-3mm}
\paragraph{Large-Scale NeRF.} To enhance the representation ability of NeRF on outdoor scenes, Mip-NeRF~\cite{DBLP:conf/iccv/BarronMTHMS21} proposed to use Gaussian to approximate the cone sampling, the integrated positional encodings are therefore scale-aware and can be used to address the 
aliasing issue of NeRF. Mip-NeRF360~\cite{DBLP:conf/cvpr/BarronMVSH22} further uses space contraction to model unbounded scenes. 
Zip-NeRF~\cite{DBLP:journals/corr/abs-2304-06706} adopted a hexagonal sampling strategy to handle the aliasing issue for 
Instant-NGP~\cite{DBLP:journals/tog/MullerESK22}. NeRF-W~\cite{DBLP:conf/cvpr/Martin-BruallaR21} models the scene changes and transient 
effects by attaching an appearance encoding for each image. Block-NeRF~\cite{DBLP:conf/cvpr/TancikCYPMSBK22} and 
Mega-NeRF~\cite{DBLP:conf/cvpr/TurkiRS22} adopt a similar divide-and-conquer strategy as 
\cite{DBLP:conf/cvpr/ZhuZZSFTQ18,DBLP:journals/pr/ChenSCW20}, where smaller blocks can be reconstructed in parallel. To this end, Block-NeRF 
focus on fixing the appearance inconsistency issue between different blocks, while Mega-NeRF aims at encouraging the sparsity of the 
network under aerial scenes. Urban-NeRF~\cite{DBLP:conf/cvpr/RematasLSBTFF22} leverages lidar points to supervise the depth of 
NeRF in outdoor scenes. SUDS~\cite{DBLP:conf/cvpr/Turki0FR23} further extended Mega-NeRF into dynamic scenes. 
Different to previous large-scale NeRF methods, Switch-NeRF~\cite{DBLP:conf/iclr/Mi023} uses a switch transformer that learns 
to assign rays to different blocks during training. Grid-NeRF~\cite{DBLP:conf/cvpr/XuXPPZT0L23} designed a two-branch network architecture, where the NeRF branch can encourage the 
feature plane~\cite{DBLP:conf/eccv/ChenXGYS22} branch recover more scene details under large-scale scenes. However, the two-branch training scheme is trivial and needs a long time to train. NeRF2NeRF~\cite{DBLP:journals/corr/abs-2211-01600} and 
DReg-NeRF~\cite{Chen_2023_ICCV} assumes images are only available during training in each block, and they propose methods to register 
NeRF blocks together.

%% file: sec/3_method.tex
\section{Our Method}
\label{sec:method}

\begin{figure*}[htbp]
    \centering
    \includegraphics[width=0.98\linewidth]{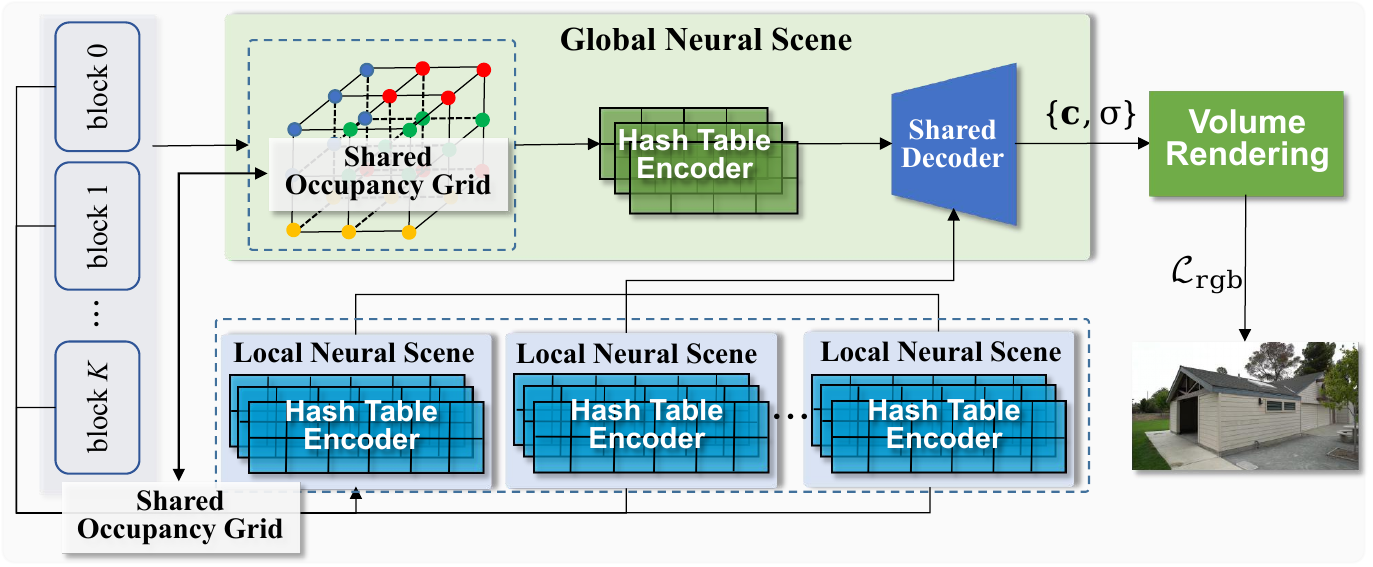}
    \vspace{-3mm}
 
    \caption{\textbf{Network architecture of our method}. Given the full set of training images, we first train 
        a coarse global model. Subsequently, we split images into different blocks. We further create overlapping regions across different 
        blocks by scaling up the bounding boxes which cover all images in local blocks. 
        We then share the decoder from the global 
        model across all local blocks, and finetuning it to better condition the hash encodings in local blocks. The final rendered 
        images are fused by a `winner-takes-all' strategy.
    }
   \label{fig:network_architecture}
   \vspace{-4mm}
\end{figure*}

Fig.~\ref{fig:network_architecture} shows an illustration of our network architecture.
Given the full set of training images, we first train 
a coarse global model. Subsequently, we split the images into different blocks. We further create overlapping regions across different blocks by 
scaling up the bounding boxes which cover all images in local blocks. 
We then share the decoder from the global model across all local 
blocks, and finetuning it to better condition the hash encodings in local blocks. The final rendered images are fused from different 
blocks with the global model as guidance.

\subsection{Preliminaries}

\paragraph{Neural Radiance Fields (NeRF).}
Neural Radiance Fields (NeRF), proposed by Mildenhall \etal~\cite{DBLP:conf/eccv/MildenhallSTBRN20} in 2020, revolutionized 3D scene 
reconstruction and novel view synthesis. 
NeRF represents scenes as a continuous function, estimating the volumetric scene representation by predicting the volume density and 
view-dependent emitted radiance at any given point. 
As introduced by NeRF~\cite{DBLP:conf/eccv/MildenhallSTBRN20}, the fundamental equation for volume rendering can be defined as:
\begin{equation}
\begin{small}
    C(x) = \int_{t_{\text{near}}}^{t_{\text{far}}} T(t) \cdot \sigma(t) \cdot L_i(t)\ dt ,
\end{small}
\end{equation}
where $C(x)$ denotes the color of a pixel, $\sigma(t)$ is the volume density, 
$L_i(t)$ is the emitted radiance, and $t_{\text{near}}$ and $t_{\text{far}}$ indicate the start and end points of the ray traversing 
the scene, respectively. $T(t)$ is the accumulated transmittance:
\begin{equation}
T(t) = \exp\left(-\int_{t_{\text{near}}}^t \sigma(s) \, ds\right).
\end{equation}

\paragraph{Instant Neural Graphics Primitives for NeRF.}
As an extension of NeRF introduced by M\"{u}ller \etal~\cite{DBLP:journals/tog/MullerESK22}, Instant Neural Graphics Primitives (InstantNGP) offers a solution to the computational challenges associated with NeRF where training time is significantly reduced from days to minutes by encoding scene representations.
By addressing hash collisions implicitly, 
InstantNGP enables faster and more memory-efficient neural scene reconstruction, offering a practical solution for large-scale scene 
reconstruction with reduced computational demands.

In this work, we utilize InstantNGP as our NeRF backbone, where the multi-resolution hash table is formulated as an encoder:
\begin{equation}
\begin{small}
    \mathbf{F} = \text{Encoder} (\mathbf{x}),
\end{small}
\end{equation}
where $\mathbf{F}$ is the feature embedding concatenated from the multi-resolution hash table 
$\mathbf{F}=[\mathbf{f}_1, \cdots, \mathbf{f}_L] \in \mathbb{R}^{L\cdot F}$, $L$ is the number of levels, and $F$ is the feature dimension.
The embedding is then decoded into volume density $\sigma$ and per-point color $\mathbf{c}$ by a shallow layer MLP:
\begin{equation}
    \sigma, \mathbf{c} = \text{Decoder} (\mathbf{F}).
\end{equation}
For unbounded scenes, we contract the 3D point $\mathbf{x}$ using the space contraction method of 
Mip-NeRF360~\cite{DBLP:conf/cvpr/BarronMVSH22}:
\begin{equation}
    \text{contract}(\mathbf{x}) = 
    \left\{
    \begin{aligned}
        \mathbf{x},\qquad &\|\mathbf{x}\| \le 1, \\
        \left( 2 - \frac{1}{\|\mathbf{x}\|} \right) \left( \frac{\mathbf{x}}{\mathbf{x}} \right),\qquad &\|\mathbf{x}\| > 1.
    \end{aligned}
    \right.
\end{equation}

\subsection{Coarse Global Reconstruction}

Our initial step involves training a coarse global model to represent the entire scene. The hash function maps a point to its hash 
entry is defined as:
\begin{equation}
    h(\mathbf{x}) = \left({\oplus}_{i=1}^d x_i \pi_i \right) \mod T,
\end{equation}
where $T$ is total number of hash entries at each level, $\oplus$ denotes the bit-wise $\text{XOR}$ operation, and $\pi_i$ are large 
unique prime numbers. 
The global model which we termed as the \textit{"coarse model"} 
shares the same hash table size as the local models. However, since a whole scene 
contains more points $\mathbf{x}$ than a part of the scene 
and the hash table size $T$ remains constant, hash collisions are more 
likely to occur in the global model compared to a local model with the same hash table size. While Instant-NGP implicitly handles hash 
collisions during training, collided hash entries in the global model may take a weighted mean of gradients from different parts of the 
scene. This equality in contribution from various parts can degrade the quality of reconstruction in the final model.
\begin{figure}[H] 
    \centering
    \includegraphics[width=0.8\linewidth]{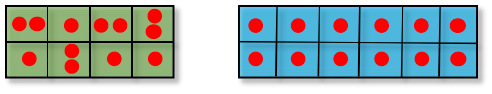}
    \vspace{-4mm}
 
    \caption{\textbf{An illustration of hash collision}. Green table: a small hash table with 8 entries; Blue table: a larger hash table 
             with 12 entries. Red circle: 3D points that are hashed into the bins.
            }
   \label{fig:hash_collision}
   \vspace{-4mm}
\end{figure}

We observe that enlarging the size of the hash table is impractical for two main reasons:
\begin{itemize}
\item \textbf{Larger model requires more memory and longer training time.} For implicit representations like MLP, a larger model demands more 
memory during training due to updates to all parameters during backpropagation. This increase in computational demands slows down training. 
For InstantNGP, although only the feature vectors of local voxels are updated, a larger memory is still required. 
Fig.~\ref{fig:hash_collision} illustrates why training a larger hash table is slower when given a fixed number of points. Suppose we have twelve 3D points 
$\{\mathbf{x}_i \mid i \in [1, N] \}$, a small hash table $\mathcal{H}_{\text{s}}$, and a larger hash table $\mathcal{H}_{\text{l}}$. 
For $\mathcal{H}_{\text{l}}$, gradients must be computed for all bins 
$[\frac{\partial \mathcal{L}}{\partial \mathbf{F}_1},\ \cdots,\ \frac{\partial \mathcal{L}}{\partial \mathbf{F}_{12}}]$. 
However, for $\mathcal{H}_{\text{s}}$, only eight bins need computing gradients during backpropagation 
$[\frac{\partial \mathcal{L}}{\partial \mathbf{F}_1},\ \cdots,\ \frac{\partial \mathcal{L}}{\partial \mathbf{F}_8}]$, as four bins contain 
more than one point due to hash collisions.

\item \textbf{Larger model does not always translate to better reconstruction quality.} This is evident in 
Fig.~\ref{fig:tat_ngp_bad_cases}, where undeterministic hash collisions can result in two parts contributing equally to the same voxels.

\end{itemize}
%

\subsection{Fine Local Reconstruction}

\paragraph{Block Splitting.} Given the complete set of training images $\{\mathbf{I}_1, \cdots, \mathbf{I}_N \}$ and their 
corresponding camera poses $\{\mathbf{P}_1, \cdots,\ \mathbf{P}_N \}$. traditional distributed Structure from Motion (SfM) 
methods~\cite{DBLP:conf/cvpr/ZhuZZSFTQ18,DBLP:journals/pr/ChenSCW20} often adopt a recursive strategy to construct intersected 
blocks. In detail, a view graph is constructed with images as nodes and the number of correspondences as edge weight. The 
graph is then split into disconnected components, and images in the current component are inserted into other components based 
on the number of correspondences. 
However, we find this strategy unsuitable for our purposes. This is because novel view synthesis in NeRF 
relies on photometric information,
but SfM primarily relies on feature-metric correspondences.
Consequently, the 
block-splitting method in SfM might yield suboptimal results, especially in texture-less regions. Additionally, distributed 
SfM necessitates splitting images before computing camera poses, while NeRF already has known camera poses 
and thus enabling us to 
leverage these priors for our block-splitting method.

We employ KMeans to cluster images close to each other into the same block: $L_{\mathbf{I}_i} = \text{KMeans} (\mathbf{t}_i)$,
where $\mathbf{t}_i$ represents camera centers, $L_{\mathbf{I}_i}$ is the block ID of image $i$.
While Mega-NeRF~\cite{DBLP:conf/cvpr/TurkiRS22} also uses KMeans to split blocks, the performance is often unsatisfactory 
due to the lack of overlapping regions between blocks.
To this end, we compute axis-aligned bounding boxes $\text{AABB}_k$ using the camera poses in each block. 
We then scale up each AABB using a scale factor vector $\mathbf{s}_{\text{AABB}}=[s_x, s_y, s_z]^{\top}$, where $s_x,s_y,s_z$ are 
the scale factor along the $x,y,z$ axis. By default, $s_x=s_y=s_z$. Next, we regroup images into different blocks, where images 
falling into the same AABB belong to the same block. This approach introduces overlapping regions between blocks since the 
AABBs intersect. By controlling $\mathbf{s}_{\text{AABB}}$, we can manage the overlapping ratio across blocks:
1) Blocks have no overlapping regions when $s=1.0$; 2) Blocks encompass all training images when $s \rightarrow \frac{\text{AABB}_{\text{whole}}}{\text{AABB}_k}$, where $\text{AABB}_{\text{whole}}$ denotes 
the AABB for the whole training images.
Our block-splitting algorithm is summarized 
in Alg.~\ref{alg:block_splitting}.

\vspace{-2mm}
\begin{algorithm}
    \caption{Block Splitting Algorithm}
    \label{alg:block_splitting}
    \begin{algorithmic}[1]
        \Require Camera poses $\mathcal{P}=\{\mathbf{P}_1, \cdots,\ \mathbf{P}_N \}$;
                 Bounding-box scale factor vector $\mathbf{s}_{\text{AABB}}=[s_x, s_y, s_z]^{\top}$;
                 number of blocks $K$.
        \Ensure Intersected blocks $\mathcal{B} = \{ \mathbf{B}_k=\{\mathbf{I}_{i,k} \} \mid k \in [1, K] \}$.
  
        \State Obtain no overlapping blocks by $\mathcal{B} \rightarrow \text{KMeans}(\mathcal{P}, K)$.
        \For{$k \in [1, K]$}
            \State Compute the bounding box for block $k$:
                \begin{equation}
                    \begin{small}
                    \text{AABB}_k = [\mathbf{A}, \mathbf{B}] = [\min (\mathcal{P}_k), \max (\mathcal{P}_k)].
                    \end{small}
                \end{equation}
            \State Compute the midpoint of the diagonal for 
                    $\text{AABB}_k$: $\mathbf{C} = (\mathbf{A} + \mathbf{B}) / 2$.
            \State Compute the ray direction of the diagonal 
                    $\mathbf{r} = (\mathbf{A} - \mathbf{B})/(\|\mathbf{A} - \mathbf{B}\|)$,
                    half of the length of the diagonal $\text{len} = \|\mathbf{B} - \mathbf{A}\|/2$.
            \State Recompute $\mathbf{A}$ and $\mathbf{B}$ by:
                \begin{align}
                    & \mathbf{A} = \mathbf{C} + \mathbf{s}_{\text{AABB}} \cdot \mathbf{r} \cdot \text{len}, \nonumber \\
                    & \mathbf{B} = \mathbf{C} - \mathbf{s}_{\text{AABB}} \cdot \mathbf{r} \cdot \text{len}.
                \end{align}

            \State Update block
                $\mathbf{B}_k \rightarrow \mathbf{B}_k + {\mathbf{I}_i},\ \forall\ \mathbf{B} \le \mathbf{P}_i \le \mathbf{A}.$
        \EndFor
    \end{algorithmic}
\end{algorithm}
\vspace{-8mm}

\paragraph{Global Conditioned Local NeRF.} With the intersected blocks now constructed, we proceed to train a local NeRF model 
for each block individually. Benefiting from fewer hash collisions, the local model can capture a higher fidelity of the scene 
block, even with a hash encoder featuring the same hash table size as the coarse global model. To further enhance the reconstruction 
quality, we adopt a strategy where the decoder from the coarse global model is shared across all local models.

The shared decoder is trained using all available training images, allowing it to effectively condition the local hash encoder and 
align the feature space of hash encodings in different local models. To optimize training speed and reduce memory footprint, each 
local model receives a copy of the shared decoder, and fine-tunes its own copy during training. This approach leverages the 
comprehensive knowledge of the shared encoder while allowing each local model to adapt to the specific details within its corresponding block.

\subsection{Final Image Blending}

Blending images from different blocks poses a non-trivial challenge as there are no explicit priors available to guide the weighting 
of image colors. 
Traditional methods such as the inverse distance weighting (IDW) scheme employed by Block-NeRF
often lead to undesired outcomes of foggy or blurry images.

\vspace{-2mm}
\begin{figure}[H]
    \centering
    \includegraphics[width=0.8\linewidth]{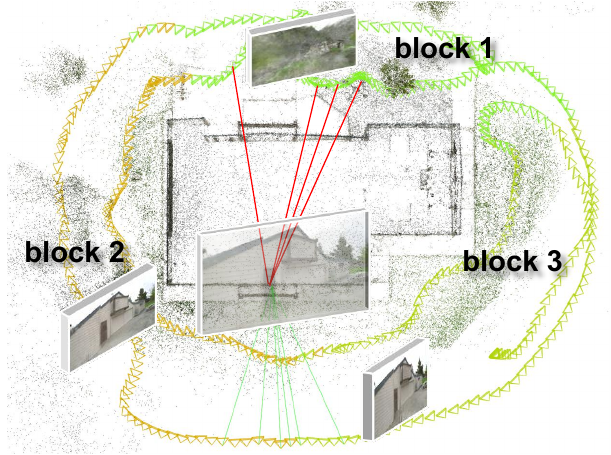}
    \vspace{-2mm}
    \caption{\textbf{Using IDW tends to generate blurry images when blocks are close to each other}. Left: The scene is split into three 
             blocks, while the point on the right-top part of the barn is only visible by two blocks. Right: Block 1 does not observe 
             that point. However, the three blocks are close to each other. IDW therefore gives approximately 
             the same weight to all blocks when blending images, which includes outliers from Block 1.}
   \label{fig:idw_blending_weakness}
   \vspace{-4mm}
\end{figure}

\vspace{-3mm}
\paragraph{Inverse Distance Weighting.} The IDW function to blend images from different blocks is defined as follows:
\begin{equation}
    \mathbf{I} = \sum_k^K \omega_k \mathbf{I}_k,\ \omega_k = d_k^{-\gamma},
    \label{eq:idw}
\end{equation}
where $d_k$ represents the distance between the query point and the block center, and $\gamma$ is a hyperparameter (default to 
$\gamma=1$ in our experiment). Despite its widespread use, the IDW approach tends to generate blurry images when blocks are in close 
proximity, as illustrated in Figure~\ref{fig:idw_blending_weakness}.

\vspace{-3mm}
\paragraph{Image Blending with Global Guidance.} An alternative solution involves computing uncertainties for the rendered color of 
each pixel and discarding results from blocks with significant uncertainties.  However, existing uncertainty computation methods 
designed for NeRF, \eg, NeRF-W~\cite{DBLP:conf/cvpr/Martin-BruallaR21} and Bayes' 
Rays~\cite{DBLP:journals/corr/abs-2309-03185}, are found to be insufficiently reliable.
Fortunately, a simple yet effective approach involves leveraging guidance from the coarse global model to fuse high-quality images. 
This is achieved by comparing the color difference between images rendered from local models and the coarse model. The coarse model 
is employed to compute the color for a point as well as local blocks. The pixel with the minimum color difference is then selected as:
\begin{equation}
    \setlength{\abovedisplayskip}{3pt}
    \setlength{\abovedisplayskip}{3pt}
    \mathbf{c} = \min (\| \mathbf{c}_k - \mathbf{c}_\text{global} \|).
\end{equation}

It is noteworthy that to maintain efficient inference, not all local models need to be considered in the comparison with the global 
model. As suggested by Block-NeRF, filtering out blocks that are highly unlikely to be observed in the current region can be achieved by training a visibility field.
However, we did not implement this filtering mechanism in our experiments since our blocks are sufficiently close.
We empirically affirm that this choice does not compromise the contribution and practicality of our method.

%% file: sec/4_experiments.tex
\section{Experiments}
\label{sec:experiments}

\paragraph{Implementations.} We use PyTorch~\cite{DBLP:conf/nips/PaszkeGMLBCKLGA19} for our implementation.
Training is performed using the Adam 
optimizer~\cite{DBLP:journals/corr/KingmaB14} for 200,000 iterations. We compare our approach with baseline methods, including vanilla 
NeRF~\cite{DBLP:conf/eccv/MildenhallSTBRN20}, Instant-NGP~\cite{DBLP:journals/tog/MullerESK22}, and our reimplementation of 
Block-NeRF~\cite{DBLP:conf/cvpr/TancikCYPMSBK22} since the original code for Block-NeRF is not publicly available. The learning rate is 
set to $5e-4$ for vanilla NeRF and $1e-3$ for all other methods. Notably, the learning rate for fine-tuning the shared decoder in our 
method is set to $5e-5$. A warm-up phase of 5,000 iterations is employed 
with the learning rate linearly increasing during this period, 
followed by exponential decay with a rate of $0.1$.
To enhance training efficiency, we incorporate the occupancy grid sampler from NeRFAcc~\cite{DBLP:journals/corr/abs-2305-04966} to skip 
empty space in our method. The occupancy grid sampler is also applied to all other methods for a fair comparison. 
It is important to 
note that the fine branch of the original vanilla NeRF is not enabled in our 
experiments due to memory constraints and increased training time.
All experiments are conducted on an NVIDIA A5000 GPU with 24GB memory. By default, the hash table size for both our coarse global model 
and local models is set to $2^{19}$. The hash tables have 16 levels of resolution, and each level features dimensions of 4. Our decoder 
has a depth of 5 with a width of 128. These hyperparameters are selected based on comprehensive experimentation and provide a good 
trade-off between performance and resource utilization.

\vspace{-3mm}
\paragraph{Datasets.} Our method is trained and evaluated on different large-scale unbounded scenes of the tanks-and-temples dataset~\cite{Knapitsch2017}. This dataset poses significant challenges as it encompasses diverse scenes with varying lighting 
conditions, and appearance changes (such as overexposure and shadows on the ground), and contains 300 to 1,200 images per scene. To 
split the dataset for training and validation, we adopt a strategy of holding out every 20 images for validation 
and utilizing the remaining images for training. 
To model appearance changes, we follow the approach of NeRF-W~\cite{DBLP:conf/cvpr/Martin-BruallaR21} by associating each image 
with an appearance embedding. The appearance embedding dimension is set to 8 for all methods. During validation, the appearance 
embeddings for the validation images are unknown. To address this, we calculate the mean of the appearance embeddings for all 
training images and employ it as the appearance embedding for the validation images. This approach ensures a consistent 
representation of appearance across the dataset during validation.

\begin{figure*}[htbp]
    \centering
    \includegraphics[width=0.95\linewidth]{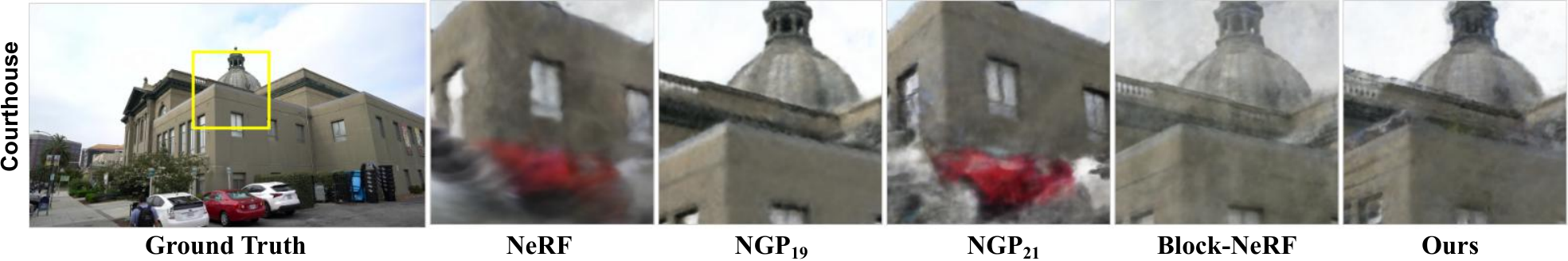}

    \vspace{-3mm}
    \caption{\textbf{Failure cases of Instant-NGP~\cite{DBLP:journals/tog/MullerESK22} on tanks-and-temples dataset~\cite{Knapitsch2017}}.
        From left to right are ground-truth image, vanilla NeRF~\cite{DBLP:conf/eccv/MildenhallSTBRN20}, 
        Instant-NGP~\cite{DBLP:journals/tog/MullerESK22} with hash table size $2^{19}$, 
        InstantNGP~\cite{DBLP:journals/tog/MullerESK22} with hash table size $2^{21}$,
        Block-NeRF~\cite{DBLP:conf/cvpr/TancikCYPMSBK22}, our Scalar-NeRF.
    }
    \vspace{-2mm}
    \label{fig:tat_ngp_bad_cases}
\end{figure*}

\begin{figure*}[htbp]
    \centering
    \subfloat {
        \includegraphics[width=0.93\linewidth]{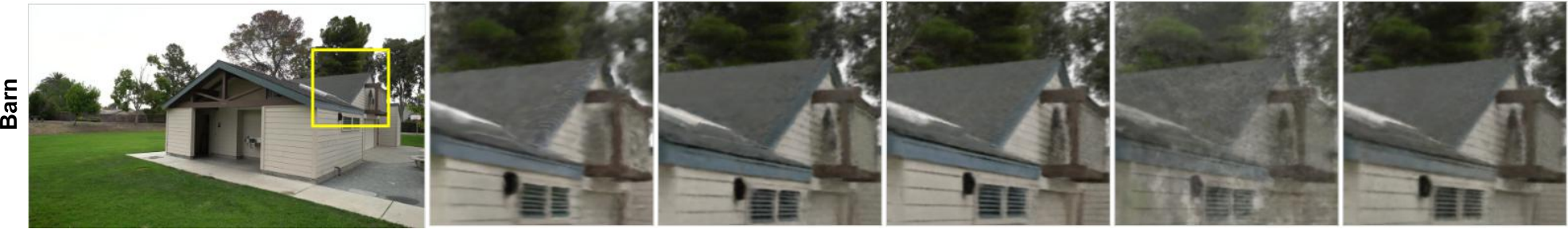}
        \label{fig:barn}
    } \\
    \subfloat {
        \includegraphics[width=0.93\linewidth]{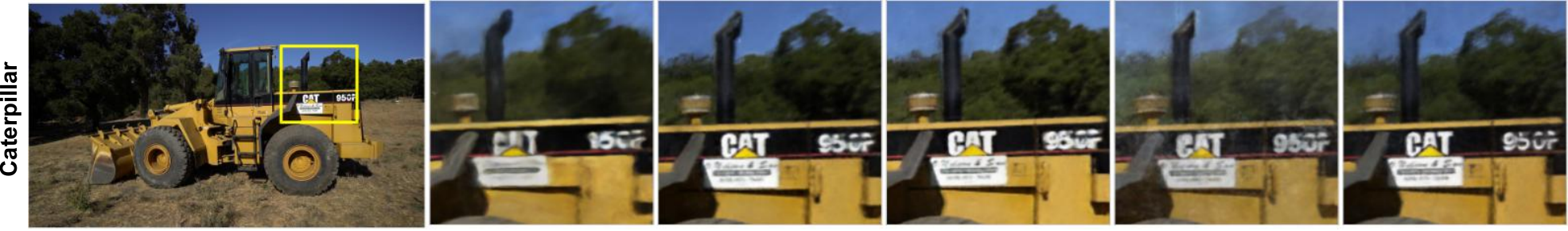}
        \label{fig:caterpillar}
    } \\
    \subfloat {
        \includegraphics[width=0.93\linewidth]{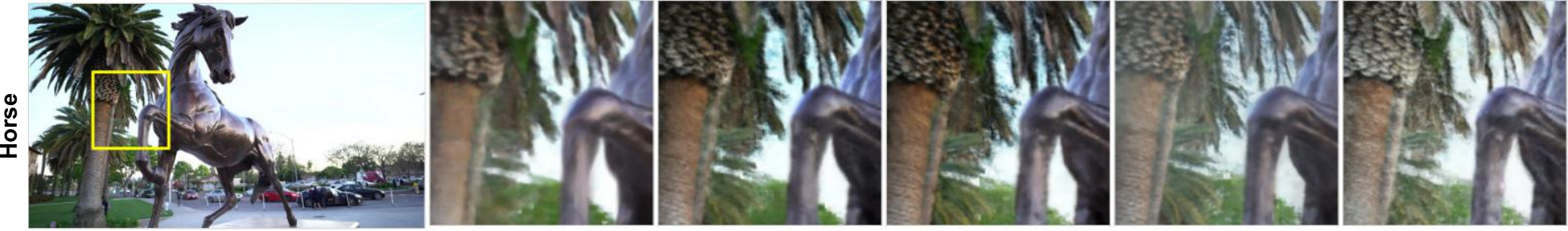}
        \label{fig:horse}
    } \\
    \subfloat {
        \includegraphics[width=0.93\linewidth]{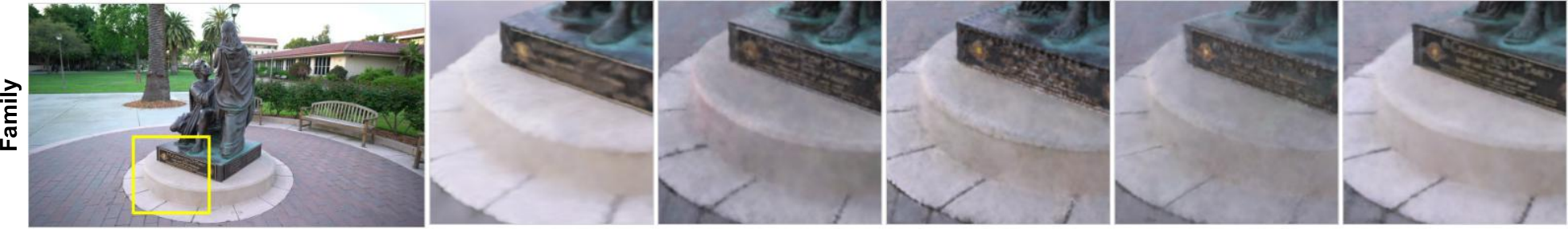}
        \label{fig:family}
    } \\
    \subfloat {
        \includegraphics[width=0.93\linewidth]{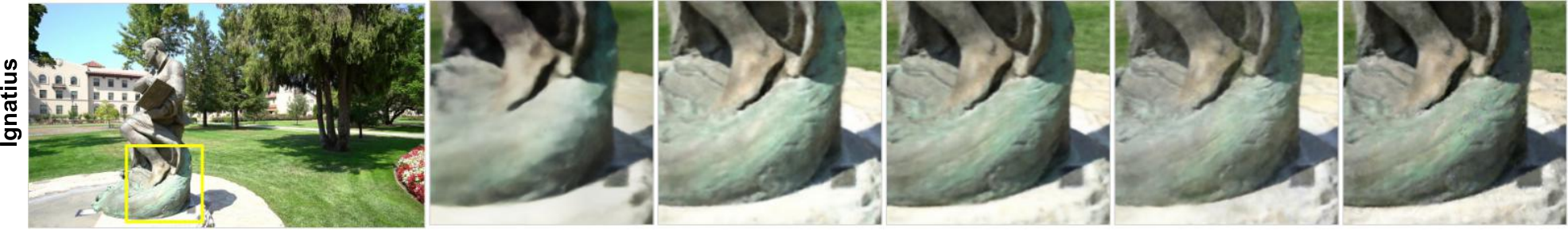}
        \label{fig:ignatius}
    } \\
    \subfloat {
        \includegraphics[width=0.93\linewidth]{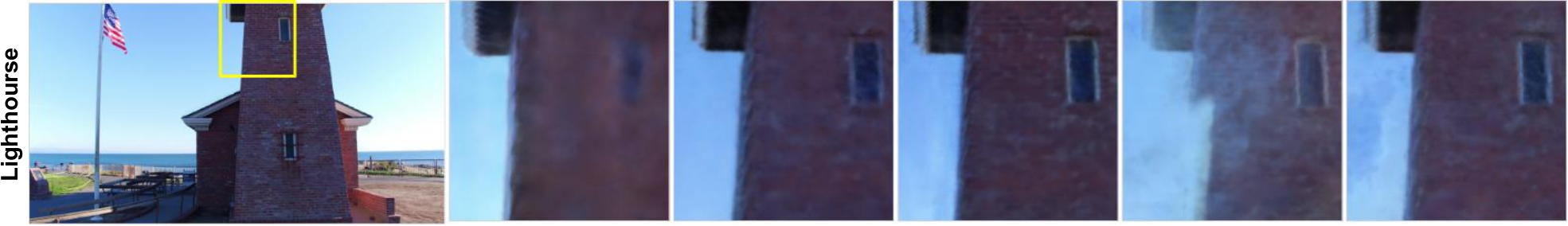}
        \label{fig:lighthouse}
    } \\
    \subfloat {
        \includegraphics[width=0.93\linewidth]{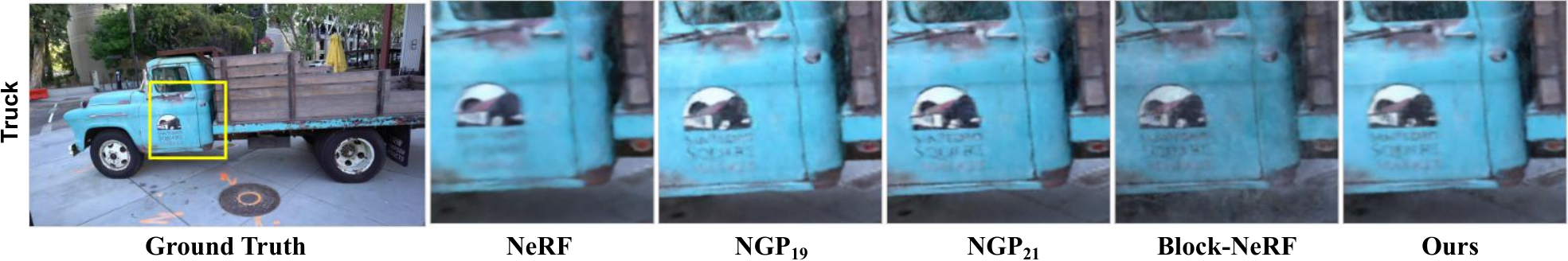}
        \label{fig:truck}
    } \\
    \vspace{-3mm}
    \caption{\textbf{The qualitative results on tanks-and-temples dataset~\cite{Knapitsch2017}}.
        From left to right are ground-truth image, vanilla NeRF~\cite{DBLP:conf/eccv/MildenhallSTBRN20}, 
        Instant-NGP~\cite{DBLP:journals/tog/MullerESK22} with hash table size $2^{19}$, 
        InstantNGP~\cite{DBLP:journals/tog/MullerESK22} with hash table size $2^{21}$,
        Block-NeRF~\cite{DBLP:conf/cvpr/TancikCYPMSBK22}, our Scalar-NeRF.
    }
    \vspace{-2mm}
    \label{fig:tat_comparison}
\end{figure*}

\vspace{-3mm}
\paragraph{Results.} We employ PSNR, SSIM~\cite{DBLP:journals/tip/WangBSS04} and LPIPS~\cite{DBLP:conf/cvpr/ZhangIESW18} as metrics for 
novel view synthesis. Quantitative results are presented in Tab.~\ref{table:tat_quantitative_nvs} 
with qualitative comparisons shown in 
Fig.~\ref{fig:tat_comparison}. From Tab.~\ref{table:tat_quantitative_nvs}, it is evident that our method excels in terms of SSIM and 
achieves the second-best result in terms of PSNR. 
Although vanilla NeRF~\cite{DBLP:conf/eccv/MildenhallSTBRN20} attains the best PSNR, it consistently produces foggy images (\cf the second column in Fig.~\ref{fig:tat_comparison}.
For InstantNGP, we provide its results with different hash table sizes of $19$ and $21$.
From Tab.~\ref{table:tat_quantitative_nvs}, we can see that $\text{NGP}_{21}$ achieves the best results in LPIPS and the second-best in SSIM. 
The reconstruction quality decreases with a reduced hash table size (\cf $\text{NGP}_{19}$ in Tab.~\ref{table:tat_quantitative_nvs}). 
Importantly, a larger hash table size does not consistently yield superior results. For the `Courthouse' scene, $\text{NGP}_{21}$ obtained 
the worst result in PSNR and performs worse than $\text{NGP}_{19}$ in terms of SSIM and LPIPS.
In Fig.~\ref{fig:tat_ngp_bad_cases}, we visualize a region where vanilla NeRF and $\text{NGP}_{21}$ inaccurately reconstructed the corner of 
the building. As expected, Block-NeRF consistently produces hazy images for almost all scenes in the tanks-and-temples dataset. Conversely, our 
method can avoid this issue by leveraging the coarse global model to discard poorly rendered results in the fusion stage. 
Our method can 
render higher-quality images than $\text{NGP}_{21}$ in details even though the hash table size is smaller than $\text{NGP}_{21}$in detailed 
areas, even with a smaller hash table size, as observed in the zoomed-in text regions of scenes `Caterpillar', `Family', and `Truck' in 
Fig.~\ref{fig:tat_comparison}. Overall, our method demonstrates superior scalability in both quantitative and qualitative evaluations compared 
to other methods.

\begin{table*}[h]
    \centering
    \resizebox{1.0\textwidth}{!}{
      \begin{tabular}{l | l l l l l | l l l l l | l l l l l }
        \toprule
  
        \multirow{2}{*}{Scenes}  &
        \multicolumn{5}{c|}{\textbf{PSNR}\ $\uparrow$} &
        \multicolumn{5}{c|}{\textbf{SSIM}\ $\uparrow$} &  
        \multicolumn{5}{c}{\textbf{LPIPS}\ $\downarrow$} \\
        
        \cmidrule(r){2-6} \cmidrule(r){7-11} \cmidrule(r){12-16}

        & \multirow{1}{*}{NeRF}
        & \multirow{1}{*}{$\text{NGP}_{19}$} 
        & \multirow{1}{*}{$\text{NGP}_{21}$} 
        & \multirow{1}{*}{Block-NeRF} & \multirow{1}{*}{Ours}
        
        & \multirow{1}{*}{NeRF}
        & \multirow{1}{*}{$\text{NGP}_{19}$} 
        & \multirow{1}{*}{$\text{NGP}_{21}$} 
        & \multirow{1}{*}{Block-NeRF} & \multirow{1}{*}{Ours}
        
        & \multirow{1}{*}{NeRF} 
        & \multirow{1}{*}{$\text{NGP}_{19}$} 
        & \multirow{1}{*}{$\text{NGP}_{21}$} 
        & \multirow{1}{*}{Block-NeRF} & \multirow{1}{*}{Ours} \\

        \midrule

        Barn
        & 22.415 & 20.422 & 22.029 & 19.934 & 21.921
        &  0.616 &  0.570 &  0.638 &  0.631 & 0.639 
        &  0.470 &  0.396 &  0.334 &  0.361 & 0.378 \\

        Caterpillar
        & 19.955 & 17.220 & 18.027 & 18.117 & 18.280
        &  0.503 &  0.506 &  0.516 &  0.516 &  0.541
        &  0.540 &  0.413 &  0.381 &  0.399 &  0.420 \\
        
        Courthouse
        & 13.073 & 15.088 & 12.586 & 15.010 & 15.990
        &  0.374 &  0.504 &  0.428 &  0.486 & 0.500
        &  0.708 &  0.553 &  0.660 &  0.590 & 0.534 \\

        Family
        & 18.832 & 17.606 & 17.197 & 16.618 & 17.838
        &  0.542 &  0.516 &  0.506 &  0.485 &  0.471
        &  0.426 &  0.331 &  0.307 &  0.362 &  0.380 \\

        Horse
        & 20.470 & 17.632 & 17.917 & 17.656 & 18.424 
        &  0.686 &  0.633 &  0.622 &  0.635 &  0.648 
        &  0.333 &  0.294 &  0.273 &  0.329 &  0.338  \\
        
        Ignatius
        & 18.217 & 15.709 & 17.285 & 16.613 & 17.498 
        &  0.431 &  0.375 &  0.481 &  0.367 &  0.458
        &  0.548 &  0.416 &  0.357 &  0.380 &  0.405 \\

        Lighthouse
        & 17.386 & 15.369 & 16.476 & 16.080 & 16.280 
        &  0.571 &  0.536 &  0.558 &  0.569 &  0.548 
        &  0.517 &  0.506 &  0.479 &  0.494 &  0.537  \\
        
        Truck
        & 17.788 & 18.562 & 18.288 & 17.722 & 17.984  
        &  0.528 &  0.575 &  0.569 &  0.556 &  0.527  
        &  0.470 &  0.356 &  0.310 &  0.341 &  0.387  \\

        \hline

        Average
        & 18.517 \mycircle[gold] & 17.201 & 17.476 \mycircle[bronze] & 17.219 & 18.027 \mycircle[silver]
        &  0.531 &  0.527 & 0.540 \mycircle[silver] &  0.531 &  0.542 \mycircle[gold] 
        &  0.512 &  0.408 \mycircle[bronze] & 0.388 \mycircle[gold]  &  0.407 \mycircle[silver] &  0.422  \\

        \bottomrule
      \end{tabular}
    }
    \vspace{-3mm}
    \caption{\textbf{Quantitative results of novel view synthesis on tanks-and-temples~\cite{Knapitsch2017} dataset}.
    $\text{NGP}_{19}$ and $\text{NGP}_{21}$ 
    denote InstantNGP~\cite{DBLP:journals/tog/MullerESK22} with 
    hash table size $2^{19}$ and $2^{21}$, respectively.
    \mycircle \mycircle[silver] and \mycircle[bronze] respectively denotes the \textcolor{gold}{first}, 
    \textcolor{silver}{second}, and \textcolor{bronze}{third} -best results.
    }
    \vspace{-5mm}
    \label{table:tat_quantitative_nvs}
\end{table*}

\vspace{-5mm}

\paragraph{Ablation Study of Network Architecture.} 
We present the ablation study of our network architecture in Tab.~\ref{table:tat_ab_study_architecture}. 
The results are averaged from the 8 scenes of the tanks-and-temples dataset.
The AABB scale factor is $[1.0,1.0,1.0]$. Specifically, `w.o.~SD' denotes our method without the shared decoder,
`w.o.~FT' denotes our method without finetuning the shared decoder for each local model,
`Full (idw)' denotes our method with inverse distance weighting function for final fusion,
`Full (global)' denotes our method with the coarse global model for final fusion. As observed, the shared decoder 
plays a critical role in our network architecture. Moreover, the local model without finetuning the shared decoder can already 
generate satisfactory results. It suggests that our network architecture has the potential to train larger models 
by increasing the hash table size and fixing the shared decoder in each local model, eliminating the need to save the 
computational graph for the parameters of the shared decoder during training.
The results indicate that our method with the IDW function performs worse than using the 
results from the coarse global model, revealing that the IDW function is not suitable when blocks are 
close to each other. Finally, our method with the full pipeline achieves the best results.
\vspace{-3mm}

\begin{table}[H]
    \centering
    \resizebox{0.45\textwidth}{!}{
      \begin{tabular}{l | r r r r r }
        \toprule
  
        \multicolumn{1}{c|}{}  &
        \multicolumn{1}{c|}{\textbf{w.o. SD}} &  
        \multicolumn{1}{c|}{\textbf{w.o. FT}} &
        \multicolumn{1}{c|}{\textbf{Full (idw)}} &
        \multicolumn{1}{c}{\textbf{Full (global)}} \\
        

    \midrule
  
        PSNR\  $\uparrow$   & 15.384 & 17.069 & 17.167 & 17.512 \mycircle[gold] \\

        SSIM\  $\uparrow$   &  0.490 &  0.537 &  0.549 &  0.565 \mycircle[gold] \\
        
        LPIPS\ $\downarrow$ &  0.521 &  0.451 &  0.437 &  0.435 \mycircle[gold] \\

        \bottomrule
      \end{tabular}
    }
    \vspace{-2mm}
    \caption{\textbf{Ablation studies of our network architecture.} `w.o. SD' denotes our method without shared decoder,
        `w.o. FT' denotes our method without finetuning the shared decoder for each local model,
        `Full (idw)' denotes our method with inverse distance weighting function for final fusion.
        `Full (global)' denotes our method with the coarse global model for final fusion.}
    \label{table:tat_ab_study_architecture}
    \vspace{-5mm}
\end{table}

\begin{figure}[H]
    \centering
    \includegraphics[width=0.99\linewidth]{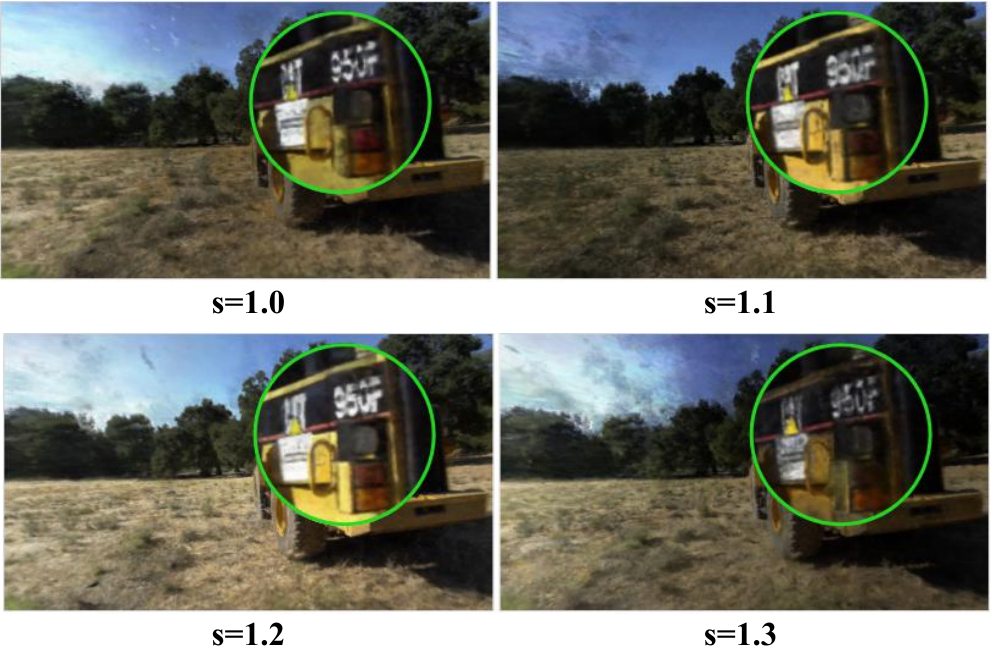}
    \vspace{-3mm}
    \caption{Qualitative results of the \textbf{ablation study for the AABB scale factor} on the `Caterpillar' scene of tanks-and-temples dataset.}
    \label{fig:caterpillar_ab_sf}
    \vspace{-2mm}
\end{figure}

\paragraph{Ablation Study of AABB Scale Factor.} 
We present the ablation study of the AABB scale factor in Tab.~\ref{table:tat_ab_study_scale_factor}. The scale factor controls the overlap regions 
across different blocks. It is expected that when we increase the scale factor, the reconstruction quality can improve since each block can 
have more observations. However, a too-large scale factor can damage the reconstruction quality. As we can see from the last column in 
Tab.~\ref{table:tat_ab_study_scale_factor}, the PSNR, SSIM, LPIPS when $s_{\text{AABB}}=1.3$ are worse than those when $s_{\text{AABB}}=1.2$.
This suggests that a moderate scale factor contributes to improved reconstruction quality, while a too-large scale factor 
may lead to a higher probability of hash collisions and a decrease in performance. We also present the 
qualitative results of the ablation study on the `Caterpillar' scene in Fig.~\ref{fig:caterpillar_ab_sf}.

\begin{table}[H]
    \centering
    \resizebox{0.48\textwidth}{!}{
      \begin{tabular}{l | r r r r r }
        \toprule
  
        \multicolumn{1}{c|}{}  &

        \multicolumn{1}{c|}{\textbf{$s_{\text{AABB}}=1.0$}} &

        \multicolumn{1}{c|}{\textbf{$s_{\text{AABB}}=1.1$}} &  

        \multicolumn{1}{c|}{\textbf{$s_{\text{AABB}}=1.2$}} &

        \multicolumn{1}{c|}{\textbf{$s_{\text{AABB}}=1.3$}} \\

        \midrule
  
        PSNR\  $\uparrow$   & 17.512 & 17.519 & \mycircle 18.146  & 17.564  \\

        SSIM\  $\uparrow$   &  \mycircle 0.565 &  0.534 &  0.555 &  0.532  \\
        
        LPIPS\ $\downarrow$ &  0.435 &  0.429 &  \mycircle 0.406 &  0.428  \\

        \bottomrule
      \end{tabular}
    }
    \vspace{-3mm}
    \caption{\textbf{Ablation studies of the AABB scale factor $s_{\text{AABB}}$.}}
    \label{table:tat_ab_study_scale_factor}
\end{table}

%% file: sec/5_conclusion.tex
\section{Conclusion}
\label{sec:conclusion}

We introduced SCALAR-NeRF, a pioneering coarse-to-fine framework designed for the intricate task of large-scale 
scene reconstruction. Utilizing the encoder-decoder architecture of NeRF, we devised a strategy to decompose expansive scenes into 
manageable blocks. Our innovation lies in the shared decoder, a component inherited by local blocks from a robust coarse global model 
and trained comprehensively on all available images. This shared decoding mechanism, coupled with the strategic use of the global model in 
the final fusion step, elegantly addresses challenges posed by imperfect inverse distance weighting functions 
as observed in approaches like Block-NeRF.
The evaluation of our SCALAR-NeRF on the demanding tanks-and-temples dataset reveals a commendable scalability in reconstruction quality. 
By leveraging the power of a global model and strategically employing it in the fusion stage, we effectively mitigate issues associated 
with foggy reconstructions.

%% file: sec/X_suppl.tex
\clearpage
\setcounter{page}{1}
\maketitlesupplementary

\section{Ablation Study of Coarse Global Model}
Our method is evaluated on the large-scale tanks-and-temples dataset. We do not evaluate our method on the city-scale scenes as in Block-NeRF due 
to GPU memory limitation, since it still requires larger memory for extremely large scenes to produce satisfactory results. The main potential 
limitation of our method is the reliance on a coarse global model. On larger scenes, the performance of the coarse global model can degrade. 
To assess the robustness and scalability of our method, we conducted an ablation study on the hash table size of the coarse global model. 
Given the inherent GPU memory limitations for extremely large scenes, we simulated scenarios where the coarse global model is trained on 
larger-scale scenes with a reduced hash table size. Specifically, we varied the hash table size of the coarse global model to $17$ and $18$, 
while maintaining the hash table size of each local model at $19$, as detailed in the main paper.

We present the quantitative results in Table~\ref{table:ablation_study_hash_size} and showcase the performance of our method under different hash 
table sizes for the coarse global model. Notably, even when the hash table size is reduced, our method consistently outperforms InstantNGP with 
an equivalent hash table size. More intriguingly, $\text{Ours}{18}$ surpasses the performance of $\text{NGP}{19}$, underscoring the potential of 
our method to scale up to larger scenes. This observation suggests that our approach maintains efficacy and competitiveness, even when confronted 
with reduced hash table sizes for the global model, signifying its adaptability and promising scalability.

\begin{table}[h]
    \centering
    \resizebox{0.48\textwidth}{!}{
      \begin{tabular}{l | l l l l l l }
        \toprule
  
        \multirow{1}{*}{Scenes} 
        & \multirow{1}{*}{$\text{NGP}_{17}$} 
        & \multirow{1}{*}{$\text{NGP}_{18}$} 
        & \multirow{1}{*}{$\text{NGP}_{19}$} 
        & \multirow{1}{*}{$\text{Ours}_{17}$} 
        & \multirow{1}{*}{$\text{Ours}_{18}$} 
        & \multirow{1}{*}{$\text{Ours}_{19}$} \\

        \midrule

        Barn
        & 19.033 & 20.461 & 20.422 & 21.733 & 22.025 & 21.921 
        \\

        Caterpillar
        & 15.709 & 17.944 & 17.220 & 15.440 & 18.509 & 18.280 
        \\
        
        Courthouse
        & 16.410 & 15.162 & 15.088 & 17.522 & 15.307 & 15.990 
        \\

        Family
        & 18.500 & 17.494 & 17.606 & 17.985 & 17.931 & 17.838 
        \\

        Horse
        & 17.304 & 18.879 & 17.632 & 17.033 & 17.376 & 18.424  
        \\
        
        Ignatius
        & 15.904 & 15.956 & 15.709 & 16.643 & 17.029 & 17.498  
        \\

        Lighthouse
        & 14.934 & 16.163 & 15.369 & 15.030 & 15.711 & 16.280  
        \\
        
        Truck
        & 18.816 & 17.706 & 18.562 & 17.495 & 17.623 & 17.984 
        \\

        \hline

        Average
        & 17.076 & 17.526 \mycircle[bronze] & 17.201 & 17.360 & 17.689 \mycircle[silver] & 18.027 \mycircle[gold]
        \\

        \bottomrule
      \end{tabular}
    }
    \caption{\textbf{Ablation study of the hash table size on tanks-and-temples~\cite{Knapitsch2017} dataset}.
    $\text{NGP}_{17}$, $\text{NGP}_{18}$ and $\text{NGP}_{19}$ 
    denote InstantNGP~\cite{DBLP:journals/tog/MullerESK22} with 
    hash table size $2^{17}$, $2^{18}$ and $2^{19}$, respectively.
    $\text{Ours}_{17}$, $\text{Ours}_{18}$ and $\text{Ours}_{19}$ 
    denote the hash table size of our coarse global model are $2^{17}$, $2^{18}$ and $2^{19}$, respectively.
    \mycircle \mycircle[silver] and \mycircle[bronze] respectively denotes the \textcolor{gold}{first}, 
    \textcolor{silver}{second}, and \textcolor{bronze}{third} -best results.
    }
    \label{table:ablation_study_hash_size}
\end{table}

\section{Uncertainty-based Image Blending}
\label{sec:uncertatin_image_blending}

In this section, we also studied the final image blending by leveraging the NeRF uncertainty. The uncertainty performs as a metric that 
indicates whether the rendered pixel color is reliable or not. Bays' rays~\cite{DBLP:journals/corr/abs-2309-03185} compute the uncertainties 
of the NeRF model after training to remove floaters. The intuition behind the method is that by adding a small perturbance to the point 
coordinate $\mathbf{x}$, the underlying geometry of the learned model should not change too much. They therefore adopt a deformation field 
to do this:
\begin{equation}
    \mathcal{D}_{\mathbf{\theta}(\mathbf{x})} = \text{Trilinear}(\mathbf{x}, \mathbf{\theta}).
\end{equation}

The volume density and per-pixel color are therefore updated as:
\begin{align}
    \sigma, \mathbf{c} = \text{Decoder}\big(\text{Encoder}(\mathbf{x} + \mathcal{D}_{\mathbf{\theta}}(\mathbf{x}))\big).
\end{align}

The uncertainty is then computed by the inverse of the diagonal components of the approximated hessian:
\begin{equation}
    \mathbf{\Sigma} \approx 
    \text{diag} \big( 
        \frac{2}{R} \sum_{\mathbf{r}} \mathbf{J}_{\mathbf{\theta}} (\mathbf{r})^{\top} \mathbf{J}_{\mathbf{\theta}} (\mathbf{r})
        + 
        2 \lambda \mathbf{I}
    \big)^{-1},
\end{equation}
where $R$ is the total number of rays $\mathbf{r}$, $\mathbf{J}_{\mathbf{\theta}}$ is the Jacobian matrix of the rendered color w.r.t. 
the model parameters $\mathbf{\theta}$.

\begin{figure*}[htbp]
    \centering
    \subfloat[Barn] {
        \includegraphics[width=0.93\linewidth]{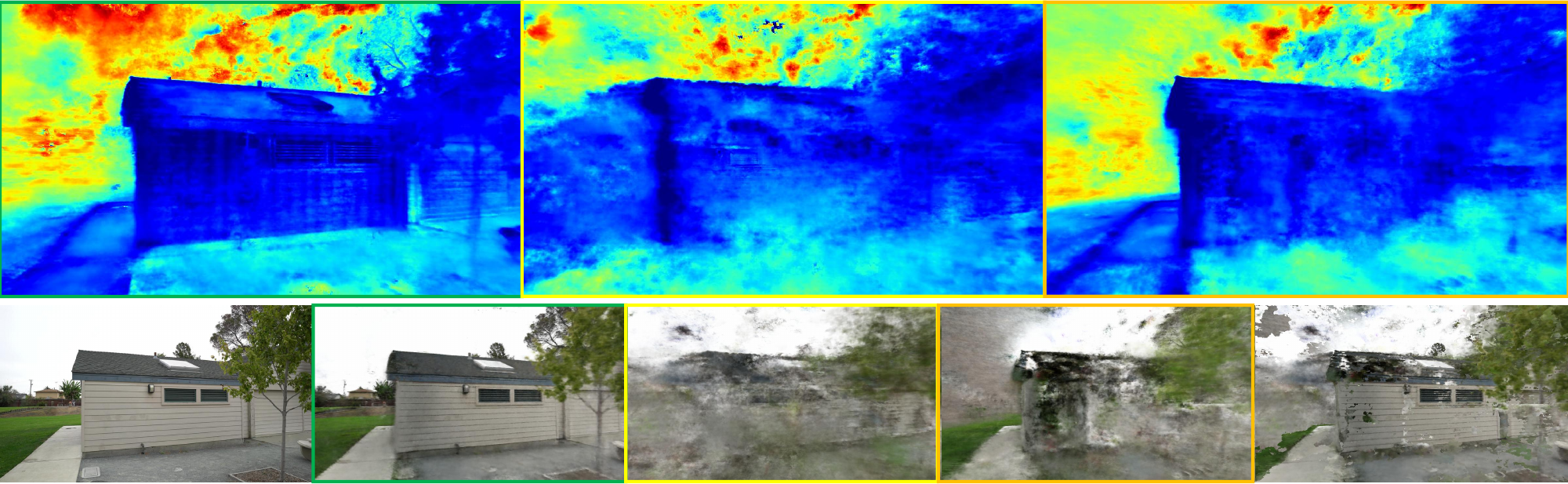}
        \label{fig:barn}
    } \\
    \subfloat[Caterpillar] {
        \includegraphics[width=0.93\linewidth]{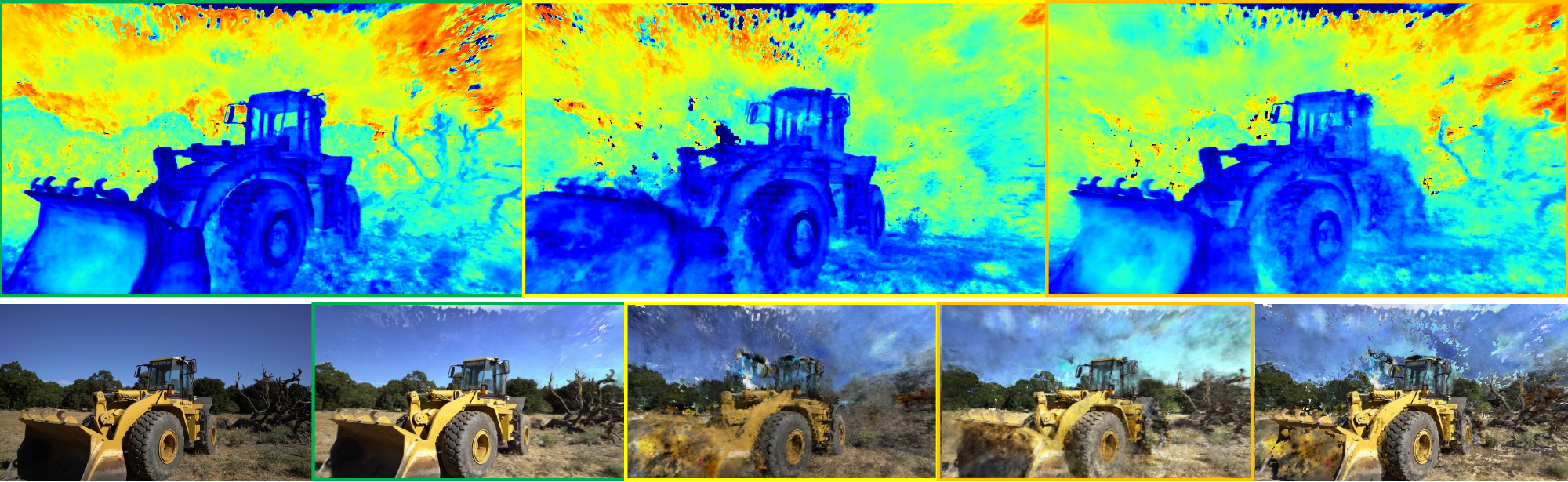}
        \label{fig:caterpillar}
    } \\
    \subfloat[Truck] {
        \includegraphics[width=0.93\linewidth]{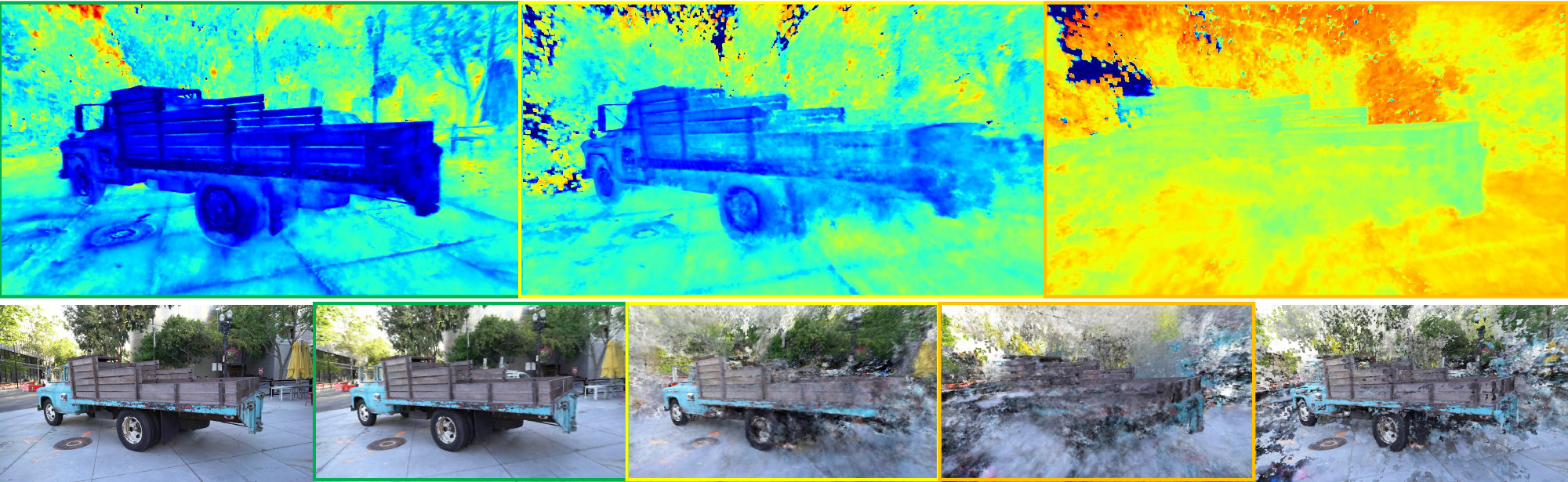}
        \label{fig:horse}
    } \\
    \caption{\textbf{The qualitative results of uncertainty-based image blending on tanks-and-temples dataset~\cite{Knapitsch2017}}.
        For each sub-figure, the first row represents the uncertainty heatmap for each block, the second row represents the ground-truth 
        image, the images rendered from each block, and the final fused images.
    }
    \label{fig:uncertainty_image_blending}
\end{figure*}

Once we computed the uncertainty for the rendered pixel, we can naturally pick out the one that has the least uncertainty for image blending.
Though Bayes' rays has shown promising results in removing floaters, we find the uncertainties are not reliable enough for high-fidelity image blending. As we can see in Fig.~\ref{fig:uncertainty_image_blending}, the uncertainty-based method can also give 
very low uncertainty for areas that are under-constrained. Moreover, for ambiguous areas, such as the sky, the uncertainty can be 
always high, which makes it difficult to filter wrong results from different blocks. The quantitative results provided on 
Tab.~\ref{table:ablation_bayes_rays} also validate the unreliability of the uncertainty-based method.

\begin{table}[H]
    \centering
    \resizebox{0.48\textwidth}{!}{
      \begin{tabular}{l | c c | c c | c c }
        \toprule
  
        \multirow{2}{*}{Scenes}  &
        \multicolumn{2}{c|}{\textbf{PSNR}\ $\uparrow$} &
        \multicolumn{2}{c|}{\textbf{SSIM}\ $\uparrow$} &  
        \multicolumn{2}{c}{\textbf{LPIPS}\ $\downarrow$} \\
        
        \cmidrule(r){2-3} \cmidrule(r){4-5} \cmidrule(r){6-7}
 
        & \multirow{1}{*}{$\text{Bayes' Rays}$} 
        & \multirow{1}{*}{$\text{Ours}$} 
        
        & \multirow{1}{*}{$\text{Bayes' Rays}$} 
        & \multirow{1}{*}{$\text{Ours}$} 
        
        & \multirow{1}{*}{$\text{Bayes' Rays}$} 
        & \multirow{1}{*}{$\text{Ours}$}  \\

        \midrule

        Barn
        & 14.517 & 21.921 
        &  0.451 &  0.639
        &  0.577 &  0.378 \\

        Caterpillar
        & 14.209 & 18.280 
        &  0.359 &  0.541
        &  0.537 &  0.420 \\
        
        Courthouse
        & 12.226 & 15.990
        &  0.419 &  0.500
        &  0.630 &  0.534 \\

        Family
        & 13.517 & 17.838 
        &  0.304 &  0.471  
        &  0.595 &  0.380 \\

        Horse
        & 13.875 & 18.424  
        &  0.410 &  0.648
        &  0.574 &  0.338 \\
        
        Ignatius
        & 12.811 & 17.498  
        &  0.235 &  0.458
        &  0.548 &  0.405 \\

        Lighthouse
        & 11.784 & 16.280  
        &  0.367 &  0.548 
        &  0.678 &  0.537 \\
        
        Truck
        & 13.223 & 17.984  
        &  0.353 &  0.527 
        &  0.543 &  0.387 \\

        \hline

        Average
        & 13.270 & 18.027
        &  0.362 &  0.542
        &  0.585 &  0.422 \\

        \bottomrule
      \end{tabular}
    }
    \caption{\textbf{Quantitative comparisons of uncertainty-based method and ours on tanks-and-temples~\cite{Knapitsch2017} dataset}.
    }
    \label{table:ablation_bayes_rays}
\end{table}

\section{Limitations and Future Work}
\label{sec:limitations}

Our approach excels in producing higher-quality images compared to Block-NeRF. However, it comes with a trade-off, as it relies on 
initializing a coarse NeRF model. Consequently, the training time for our method is marginally longer than that of Block-NeRF, especially 
when employing the Mip-NeRF backbone with InstantNGP. Additionally, during inference, our method necessitates the generation of a 
reference image from the coarse model, introducing a slight slowdown in inference speed.

While our work aims to strike a balance between computational efficiency and reconstruction quality, addressing these limitations, 
particularly in the context of larger scenes, and reducing the dependence on coarse global models presents avenues for future exploration. 
These considerations underscore the necessity for ongoing refinement and adaptation to diverse environmental scales and conditions.

\section{Social Impact}
\label{sec:social_impact}

Our method presents a significant stride forward in the field of large-scale scene reconstruction. Beyond its technical prowess, the potential 
social impact of this innovation is multifaceted.

\begin{itemize}
\item \textbf{Enhanced Visualizations}. The ability to reconstruct large-scale scenes with improved fidelity and scalability can have a profound 
impact on various industries. From urban planning to environmental monitoring, SCALAR-NeRF's capacity to generate high-quality visualizations can 
aid decision-makers and stakeholders in better understanding and analyzing complex spatial data.

\item \textbf{Accessibility and Resource Efficiency}. The scalability of SCALAR-NeRF, demonstrated by its capability to reconstruct extensive 
scenes on a single GPU, holds the promise of democratizing access to advanced reconstruction technologies. This accessibility could empower 
researchers, educators, and professionals who may have been limited by resource constraints, fostering innovation and collaboration.

\item \textbf{Advancements in Cultural Preservation}. Large-scale scene reconstruction is crucial in fields like archaeology and cultural heritage 
preservation. SCALAR-NeRF's ability to faithfully capture detailed scenes can contribute to the preservation and documentation of historical sites 
and artifacts, aiding in cultural conservation efforts globally.

\item \textbf{Improved Urban Planning}. As urban areas expand, the demand for sophisticated tools in urban planning intensifies. 
SCALAR-NeRF's detailed reconstructions can provide urban planners with realistic and data-rich representations of large-scale environments, 
facilitating more informed decision-making in areas such as infrastructure development and disaster preparedness.

\item {Training and Simulation}. In domains like autonomous vehicles and robotics, realistic scene reconstructions are invaluable for training and 
simulation. SCALAR-NeRF's ability to efficiently handle large-scale scenes can contribute to advancements in autonomous systems by providing 
realistic training environments.

\end{itemize}

While the immediate impact is within the realms of research and technology, the broader social implications lie in the democratization of 
advanced reconstruction capabilities, fostering innovation, and contributing to the understanding and preservation of our physical environment 
and cultural heritage.